 \documentclass[pmlr,twocolumn]{jmlr} 

\usepackage{booktabs}
\usepackage[load-configurations=version-1]{siunitx} 
 \usepackage{amsmath}
 \usepackage{mathtools}
\usepackage{listings}

\newcommand{\comment}[1]{}

\theorembodyfont{\upshape}
\theoremheaderfont{\scshape}
\theorempostheader{:}
\theoremsep{\newline}

\title[Uncertainty-Aware Ensembling for Severity of Alzheimer's Dementia]{Uncertainty-Aware Multi-Modal Ensembling for \titlebreak Severity Prediction of Alzheimer's Dementia}

  \author{%
   \Name{Utkarsh Sarawgi\textsuperscript{1}} \Email{utkarshs@mit.edu}\\
   \Name{Wazeer Zulfikar\textsuperscript{1}} \Email{wazeer@mit.edu}\\
   \Name{Rishab Khincha\textsuperscript{1, 2}} \Email{rkhincha@mit.edu}\\
   \Name{Pattie Maes\textsuperscript{1}} \Email{pattie@mit.edu}\\
   \addr \textbf{\textsuperscript{1}}Massachusetts Institute of Technology, 
   USA\\
   \addr \textbf{\textsuperscript{2}}BITS Pilani Goa Campus, India
  }

\begin{document}

\maketitle

\begin{abstract}
Reliability in Neural Networks (NNs) is crucial in safety-critical applications like healthcare, and uncertainty estimation is a widely researched method to highlight the confidence of NNs in deployment. In this work, we propose an uncertainty-aware boosting technique for multi-modal ensembling to predict Alzheimer's Dementia Severity. The propagation of uncertainty across acoustic, cognitive, and linguistic features produces an ensemble system robust to heteroscedasticity in the data. Weighing the different modalities based on the uncertainty estimates, we experiment on the benchmark ADReSS dataset, a subject-independent and balanced dataset, to show that our method outperforms the state-of-the-art methods while also reducing the overall entropy of the system. The source code is available at \href{https://github.com/wazeerzulfikar/alzheimers-dementia}{\lstinline|https://github.com/wazeerzulfikar/alzheimers-dementia|}
\end{abstract}


\section{Introduction}\label{sec:intro}

Alzheimer's disease is a progressive disorder that causes brain cells to degenerate and is the most common cause of dementia worldwide. It causes cognitive and behavioural deterioration of the patients \citep{molinuevo2011role} which is reflected through memory loss and language impairment \citep{escobar2010calidad}. The number of people with dementia worldwide in 2015 was estimated to be 47.47 million, and is predicted to reach 135.46 million by 2050 \citep{prince2013global}.
Recent works such as \citet{fraser2016linguistic, luz2018method, masrani2018detecting, mirheidari2018detecting, haider2019assessment, kong2019neural, pulido2020alzheimer, luz2020alzheimer, rohanianmulti, balagopalan2020bert, sarawgi2020multimodal} have taken significant steps towards more robust methods to estimate the severity of Alzheimer's Dementia (AD) using spontaneous speech and their transcriptions. However, the above methods only provide a point estimate. In such a safety-critical setting, generating a confidence measure for the neural network (NN) along with its prediction can greatly improve the reliability and awareness of the model. 

Multiple probabilistic methods, Bayesian such as \citet{graves2011practical, blundell2015varinf, hernandez2015probabilistic, kingma2015dropout, gal2016dropout, lee2017deep, wu2018deterministic, pearce2020uncertainty} and non-Bayesian such as \citet{osband2016risk, lakshminarayanan2017simple, dusenberry2020analyzing, sarawgi2020unified}, have been proposed to quantify the uncertainty estimates. In a multimodal setting, the ensembling of networks can further improve performance of models. Boosting algorithms such as Adaptive Boosting \citep{freund1997adaboost}, Gradient Boosting \citep{friedman2000gradientboosting} and XG Boosting \citep{chen2016xgboost} are widely used ensembling techniques. There has been some work on incorporating uncertainty with ensembling methods. \citet{kendall2018multitask} learnt multiple tasks by using the uncertainty predicted as weights for the losses of each of the models, thus outperforming individual models trained on each task.  \citet{chang2017activebias} uses uncertainty estimates and prefer to learn the datapoints predicted incorrectly with higher uncertainty in different mini-batches of SGD.

We propose an uncertainty-aware ensembling technique for a multimodal system where each of the base learners correspond to the different modalities. The ensemble is trained in a boosted manner such that the base learners are sequentially boosted by weighing the loss with the datapoint's predictive uncertainty quantified by the predicted standard deviation (Section \ref{sec:model_arch}). We use a non-Bayesian method to estimate the predictive uncertainty for each of the base learners. The motivation is to sequentially boost the datapoints for which a particular base learner is more uncertain about its prediction. This mechanism aims to decrease the overall entropy of the multimodal system while generating uncertainty estimates. To the best of our knowledge, we are the first to explore boosting in an ensemble using uncertainty estimates predicted by individual base learners. We apply our technique in a multi-modal setting to predict the severity of AD through Mini-Mental State Exam (MMSE) scores \citep{tombaugh1992mini}.


\section{Methods and Materials}\label{sec:methods}
\subsection{Dataset}\label{sec:dataset}
We evaluate on the standardized and benchmark ADReSS dataset \citep{luz2020alzheimer}. This dataset consists of 156 speech samples and their corresponding transcripts, each from a unique subject, matched for age and gender, and their corresponding MMSE scores as labels for regression. A standardized train-test split of around 70\%-30\% 
(108 and 48 subjects) is provided by the dataset. We further split the train set into 80\%-20\% train-val sets. The test set was held out for all experimentation until final evaluation.

\subsection{Feature Engineering} \label{sec:feature_eng}
Dementia is generally characterised by symptoms of cognitive decline and an impairment in communication, memory and thinking \citep{pulido2020alzheimer}. Various features have been used to propose speech recognition based solutions for automatic detection of mild cognitive impairments from spontaneous speech \citep{toth2018speech, pulido2020alzheimer}. We extract such relevant cognitive and acoustic features that highlight the domain knowledge and context, and feed them into their respective neural models  (Section \ref{sec:model_arch}). Upon exploring the data, we identify three feature sets - which we refer to as `Disfluency', `Acoustic', and `Interventions' - to be highly correlated to the severity of AD. They are explained in details in Appendix~\ref{apd:features}.


\subsection{Model Architecture and Training}\label{sec:model_arch}

Let $\mathbf{x} \in \mathbb{R}^d$ represent a set of $d$-dimensional input features and $y \in \mathbb{R}$ denote the label for regression.  Given a training dataset $\{(\mathbf{x}_n, y_n)\}_{n=1}^N$ consisting of N i.i.d. samples, we model the probabilistic predictive distribution $p_\theta (y|\mathbf{x})$ using NN with parameters $\theta$.

Appendix \ref{apd:modelarch} illustrates and explains the architecture of the individual disfluency, acoustic, and interventions models. Each of the three models predicts a target distribution instead of a point estimate to account for the heteroscedasticity in data and yield predictive uncertainties along with the predicted mean value. This target distribution is modelled as a Gaussian distribution $p_\theta(y_n|\mathbf{x}_n)$ parameterized by the mean $\mu_\theta$ and the standard deviation $\sigma_\theta$, predicted at the final layer of the models i.e. $y_n \thicksim \mathcal{N}\left(\mu_\theta(\mathbf{x}_n),  \sigma_\theta^2(\mathbf{x}_n)\right)$.

Each of the models is trained with their corresponding input features $\mathbf{x}$ and their ground truth labels $y$ using a proper scoring rule $l(\theta, \mathbf{x}, y)$. We optimize for the negative log-likelihood (NLL) of the joint distribution $p_\theta(\mathbf{y}_n|\mathbf{x}_n)$ according to \equationref{NLL}.
\begin{align}
    \label{NLL}
    -\log p_\theta(y_n|\mathbf{x}_n) &= \frac{\log \sigma^2_\theta(\mathbf{x})}{2} + \frac{\left(y-\mu_\theta(\mathbf{x})\right)^2}{2\sigma_\theta^2(\mathbf{x})}\nonumber\\
    &+ \text{constant}
\end{align}
Each training run used a batch size of 32; and Adam optimizer with a learning rate of 0.00125 to minimize the NLL. We use boosting to train the ensemble, wherein the three individual models are the base learners. A `vanilla ensemble' would use RMSE values to weigh the loss while sequentially boosting across the base learners, and then average the predictions of all the base learners for the final prediction. We propose an `uncertainty-aware ensemble' wherein instead of RMSE values, we use predictive uncertainty quantified by the predicted standard deviation to weigh the loss while sequentially boosting across the base learners, and average the predictions of all the base learners for the final prediction. We refer to this method as `UA ensemble'. A variation to the averaging of the predictions is also experimented. In this method, our UA ensemble performs a weighted average of the predictions, where the weights used are the inverse of the respective predictive uncertainty, quantified by the predicted standard deviation. This is shown in \equationref{eq:ua_voting}, where $P(\mathbf{x}_n)$ is the final prediction and $N$ is the total number of individual modalities. We refer to this variation as `UA ensemble (weighted)'.
\begin{align}
\begin{split}
{P(\mathbf{x}_n)} = \frac{\sum_{i=1}^N\frac{1}{\sigma_{\theta_i}(\mathbf{x}_n)}\mu_{\theta_i}(\mathbf{x}_n)}{\sum_{i=1}^N\frac{1}{\sigma_{\theta_i}(\mathbf{x}_n)}}\\
\label{eq:ua_voting}
\end{split}
\end{align}


\begin{figure*}[htb!] 
\centering 
  \begin{minipage}{.45\textwidth}
  \centering
  \includegraphics[width=0.8\linewidth]{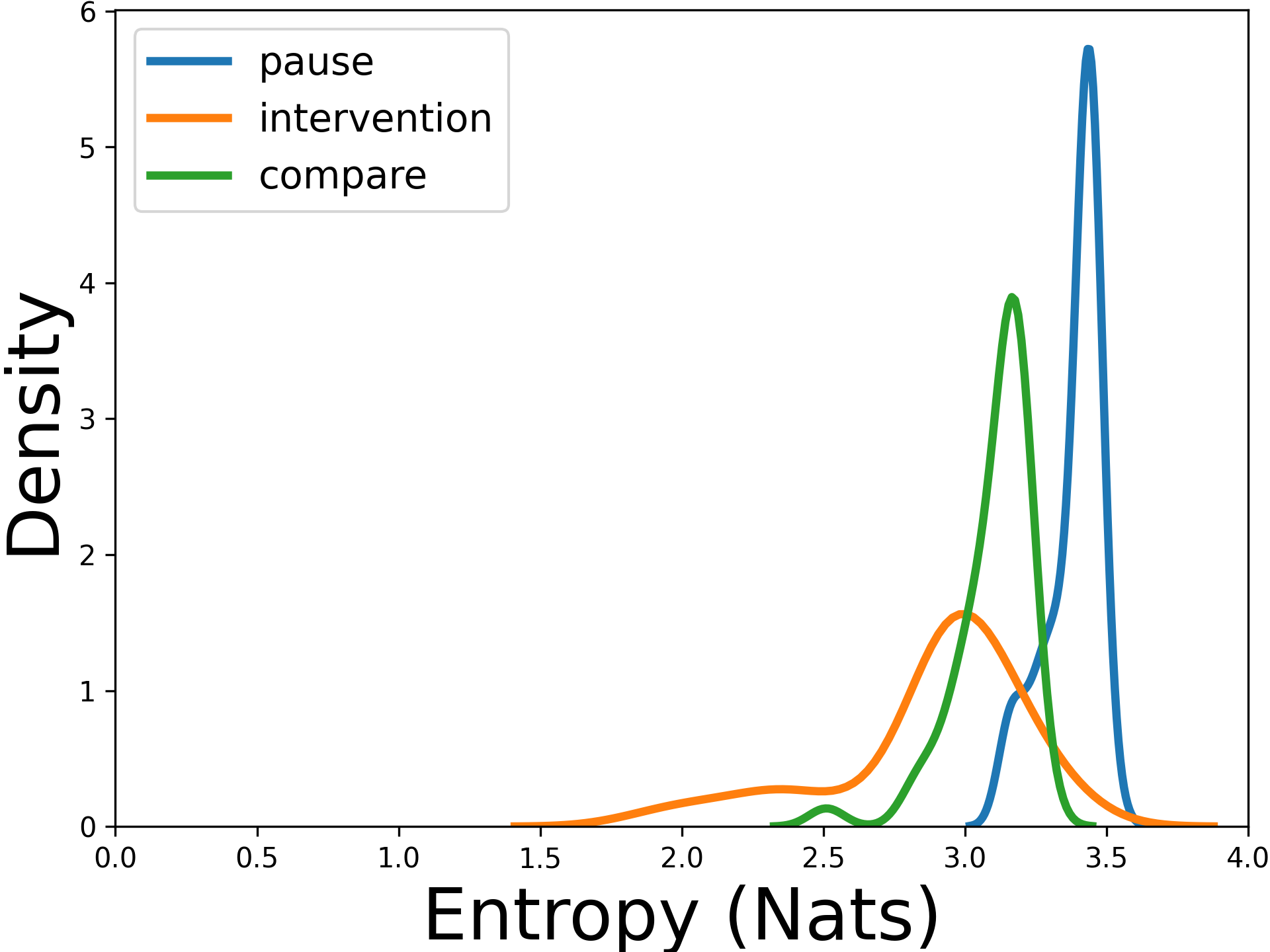}
\end{minipage}%
~
\begin{minipage}{.45\textwidth}
\centering
  \includegraphics[width=0.8\linewidth]{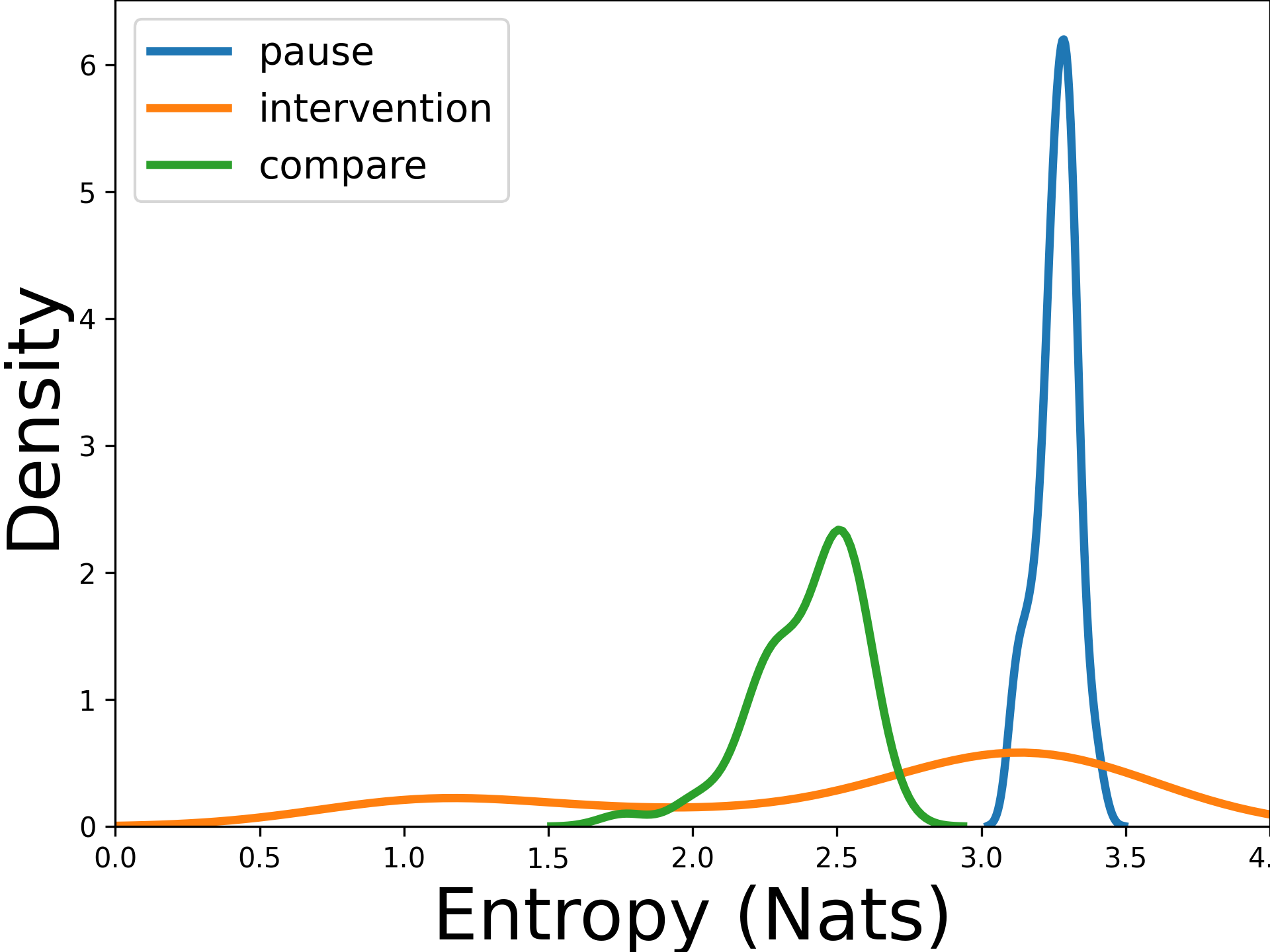}
\end{minipage}%
  \caption{Entropy analysis, using kernel density estimation plots, of the base learners in the ensemble methods while sequentially boosting using RMSE values (left) and uncertainty estimates (right) as loss weights. The decrease in overall entropy of the uncertainty-aware boosting shows the decrease in uncertainty in the predictions.}\label{fig:entropy}
\end{figure*}
\vspace{-11mm}
\section{Results}
\label{sec:results}

For robustness, we repeat every evaluation 5 times using random seeds and report the mean and the variance of the RMSE results. We first evaluate each modality i.e. base learner individually and compare with the vanilla and uncertainty-aware ensembles (\tableref{tab:res1}). The order of sequential boosting for the propagation of the uncertainties is chosen in the order of the individual performance on the test set.


\begin{table}[hbtp]
\floatconts
  {tab:res1}
  {\caption{Comparison of individual modalities i.e. base learners and ensemble methods on test set of ADReSS dataset.}\vspace{-1mm}}
  {\begin{tabular}{lc}
  \toprule
  \bfseries Model & \bfseries RMSE\\
  \midrule
  Disfluency & 5.71 $\pm$ 0.39\\
  Acoustic & 6.66 $\pm$ 0.30\\
  Interventions & 6.41 $\pm$ 0.53\\
  Vanilla Ensemble & 5.17 $\pm$ 0.27\\
  UA Ensemble & 5.05 $\pm$ 0.53\\
  UA Ensemble (weighted) & \textbf{4.96 $\pm$ 0.49}\\
  \bottomrule
  \end{tabular}}
\end{table}

The use of uncertainty awareness can improve the robustness of the ensemble with uncertain datapoints i.e subjects. To highlight this, we evaluate the entropy of the base learners in the ensemble methods while sequentially boosting in the vanilla ensemble and the UA ensemble. Upon comparison, we observe a decrease in the overall entropy of the system when the ensemble is uncertainty-aware (Figure \ref{fig:entropy}). The increased reduction in the entropy as we sequentially move from the first base learner to the last base learner of the ensemble further indicates the significance of introducing uncertainty awareness into the ensemble.

We also compare our uncertainty-aware ensemble methods with current state-of-the-art methods on the ADReSS test set. \tableref{tab:res2} shows that the best of 5 runs of UA Ensemble is competitive, and that of the UA Ensemble (weighted) outperforms other methods. 

\begin{table}[hbt!]
\setlength{\tabcolsep}{4pt}
\floatconts
  {tab:res2}
  {\caption{Comparison of uncertainty-aware ensemble methods with state-of-the-art methods on the ADReSS test set.}}
  {\begin{tabular}{lc}
  \toprule
  \bfseries Model & \bfseries RMSE\\
  \midrule
  \citet{pappagariusing} & 5.37\\
  \citet{luz2020alzheimer}   & 5.20\\
  \citet{sarawgi2020multimodal} & 4.60\\
  \citet{searle2020comparing} & 4.58\\
  \citet{balagopalan2020bert} & 4.56\\
  \citet{rohanianmulti} & 4.54\\
  \citet{sarawgi2020unified} & 4.37\\
  UA Ensemble & 4.35\\
  UA Ensemble (weighted) & \textbf{3.93}\\
  \bottomrule
  \end{tabular}}
  \vspace{-1mm}
\end{table}

\section{Discussion and Future Work}
\label{sec:discussion}

We propose an uncertainty-aware ensembling method in a safety-critical application where the reliability and awareness of an NN is important. The propagation of the uncertainty sequentially through the base learners of every modality aids the multi-modal system to decrease the overall entropy in the system, making it more reliable when compared to vanilla ensembles. Due to the use of a subject-independent dataset, the uncertainty of a datapoint can be tied to subject characteristics. By encouraging the ensemble to focus more on cases of increased uncertainty and the use of uncertainty-weighting aimed to leverage the individual uncertainties, we show how our uncertainty-aware ensemble outperforms the results of state-of-the-art methods. Further, the availability of predictive uncertainties corresponding to each modality can assist the user in making more informed decisions.

This work aims to encourage fair and aware models. Possible future directions to this work would be to experiment with the order of the modalities in the ensemble, a more advanced way to weigh the base learners, and experiment with Bayesian methods for estimating uncertainty.


\clearpage
\bibliography{jmlr-sample}

\clearpage
\appendix

\section{Feature Engineering}\label{apd:features}

The three  feature  sets  - namely  `Disfluency’,  `Acoustic’,  and `Interventions’ mentioned in Section \ref{sec:feature_eng} are explained below :

\textbullet\hspace{1mm}\textit{Disfluency:} A set of 11 distinct and carefully curated features from the transcripts, like word rate, intervention rate, and different kinds of pause rates reflecting upon speech impediments like slurring and  stuttering. These are normalized by the respective audio lengths and scaled thereafter.

\textbullet\hspace{1mm}\textit{Acoustic:} The ComParE 2013 feature set \citep{eyben2013recent} was extracted from the audio samples using the open-sourced openSMILE v2.1 toolkit, widely used for affect analyses in speech \citep{eyben2010opensmile}. This provides a total of 6,373 features that include energy, MFCC, and voicing related low-level descriptors (LLDs), and other statistical functionals. This feature set encodes changes in speech of a person and has been used as an important noninvasive marker for AD detection \citep{lopez2012alzheimer, luz2020alzheimer}. Our system standardizes this set of features using z-score normalization, and uses principal component analysis (PCA) to project the 6,373 features onto a low-dimensional space of 21 orthogonal features with highest variance. The number of orthogonal features was selected by analyzing the percentage of variance explained by each of the components.

\textbullet\hspace{1mm}\textit{Interventions:} Cognitive features reflect upon potential loss of train of thoughts and context. Our system extracts the sequence of speakers from the transcripts, categorizing it as subject or the interviewer. To accommodate for the variable length of these sequences, they are padded or truncated to length of 32 steps, found upon analyses and tuning of sequence lengths.

\section{Model Architecture}\label{apd:modelarch}

The individual model architectures used for each of the feature sets are shown in \figureref{fig:modelarch}. The disfluency model is a multi-layer perceptron (MLP) that projects the 11-feature input to a higher dimensional space for better separability of the features. The acoustic model is an MLP with a single hidden layer that adds non-linearity and regularizes the PCA (principal component analysis) decomposed feature space. The interventions model uses a recurrent architecture to learn the temporal relations from the sequence of interventions. We successively added and removed layers and tuned regularizers to reach the final architecture of the three models.
\begin{figure}[htb!] 
\floatconts
  {fig:modelarch}
  {\caption{Architecture of the (1) Disfluency (2) Acoustic and (3) Intervention models}}
  {\includegraphics[width=0.59\linewidth]{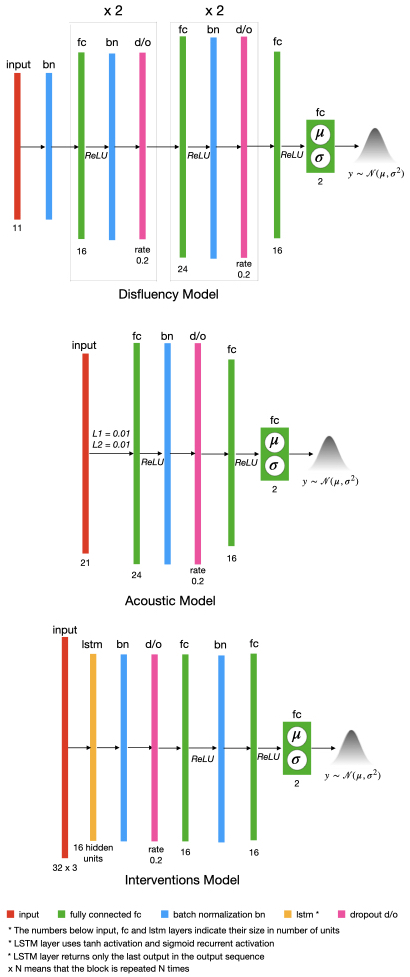}}
\end{figure}
\end{document}